\documentclass[11pt]{article} 

\usepackage[preprint]{acl}

\usepackage{times}
\usepackage{latexsym}

\usepackage[T1]{fontenc}
\usepackage{tikz}
\usepackage{amsmath}

\usetikzlibrary{shapes.geometric, arrows.meta, positioning, calc}

\usepackage[utf8]{inputenc}
\usepackage{booktabs}   
\usepackage{tabularx}   
\usepackage{array}      
\usepackage{microtype}

\usepackage{inconsolata}

\usepackage{graphicx}
\usepackage[dvipsnames]{xcolor}

\newcommand{\advplus}{\textcolor{ForestGreen}{\textbf{+}}\ }
\newcommand{\limminus}{\textcolor{BrickRed}{\textbf{--}}\ }
\title{Long-term Measurements:\\Towards a Longitudinal Understanding of Human-AI Interactions}

\author{
  \textbf{Nicole Mitchell}$^{\diamondsuit}$ \quad
  \textbf{Dhruv Agarwal}$^{\heartsuit,\spadesuit}$ \quad
  \textbf{Maty Bohacek}$^{\heartsuit,\clubsuit}$ \\
  \textbf{Remi Denton}$^{\diamondsuit}$ \quad
  \textbf{Roma Patel}$^{\heartsuit}$ \\[1.5ex]
  $^{\heartsuit}$Google DeepMind \quad
  $^{\diamondsuit}$Google Research \quad
  $^{\spadesuit}$Cornell University \quad
  $^{\clubsuit}$Stanford University \\[1ex]
  \texttt{\{nicolemitchell, dentone, romapatel\}@google.com} \\
  \texttt{da399@cornell.edu}, \texttt{maty@stanford.edu}
}

\begin{document}
\maketitle


\setcounter{footnote}{0}
\renewcommand{\thefootnote}{\arabic{footnote}}

\begin{abstract}
Language models have taken on the role of a very new type of technology, by virtue of their "human-ness" and rapid integration into users' daily lives.
This combination of features can introduce longitudinal risks---cognitive, developmental and socio-affective changes in humans---that might not surface during a short-term interaction, but can have lasting long-term effects on users.
This forms the basis of a critical new mission for NLP: to pivot from static, short-term evaluations of text generations to long-term measurements of behavioral changes, towards a diachronic understanding of human-model interactions. 
In this work, we draw from measurements used in social science fields that are crucial to understand emergent phenomena in longitudinal data. We discuss how computational methods in the field of NLP need to be combined with such measurements, not only to understand long-term safety risks of human-model interactions, but to help steer model development towards positive rather than negative outcomes for users.
This ability to model human behavioral shifts as a function of model interactions can facilitate online rather than post-hoc detection of problematic behaviors, and should be leveraged in alignment frameworks to mitigate long-term risks in users.
\end{abstract}

\section{Introduction}

For decades, the mission of NLP has been to build theories and models of human language in order to achieve human-level language understanding and generation. At this point in time, with models having largely reached this milestone,\footnote{We acknowledge the large body of work that finds systemic failures in reasoning, logic and generation of models. This statement only refers to the sufficient level of fluency that has led to their integration into daily human infrastructure, thus leading to the longitudinal risks we put forward.} the field of NLP faces a new, and more urgent mission: measuring the developmental impact of models on human users through continued interactions.

\begin{figure}[htbp]
\centering
\includegraphics[width=\linewidth]{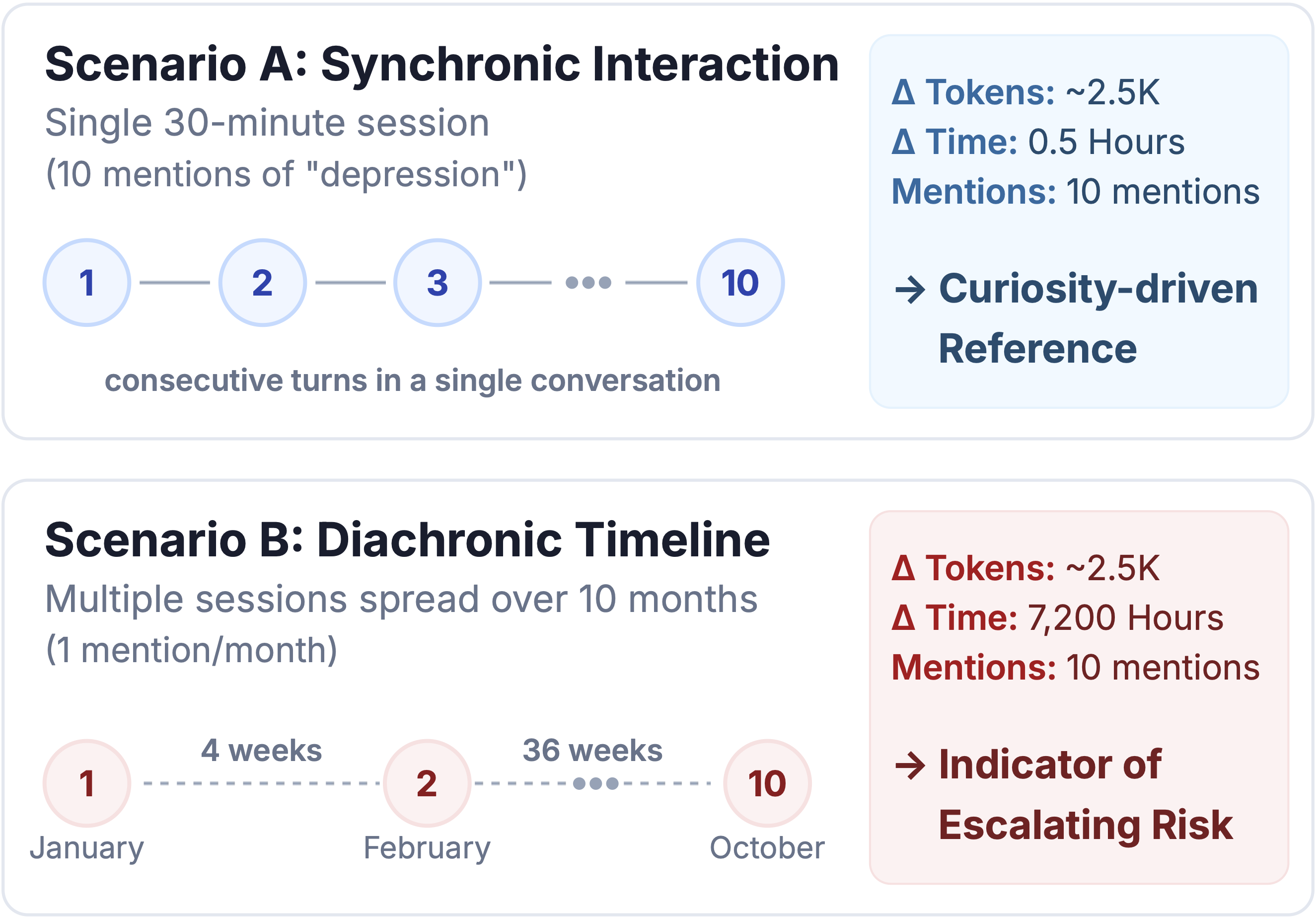}
\caption{Two scenarios where a user's reference to depressive thoughts might indicate long-term safety risks: a single synchronic interaction (Scenario \textbf{A}) vs. a diachronic trend (Scenario \textbf{B}). The longitudinal trend in \textbf{B} provides a more complete understanding of high-risk user behaviors compared to single-session\protect\footnotemark evaluations.}
\label{fig:token_time_confound}
\vspace{-0.5em}
\end{figure}
\footnotetext{We define a session to be an uninterrupted period of time in which a user converses with a chatbot.}

A large part of this urgency is in reaction to how quickly these models have been adopted---and sometimes unknowingly integrated---into users' everyday lives \cite{susser2019invisible}. This paradigm shift of seamless integration of AI into user applications brings about several \textit{longitudinal} risks i.e., ones that arise over the course of sustained interactions with systems. 
When coupled with other known risks such as anthropomorphizing AI, this can lead to over-attachment, and unnatural relationship formation with AI systems \cite{cohn2024believing, weidinger2021ethical}. This perceived "real-ness" is likely exacerbated by explicit cross-session memory, personalization and general linguistic competence of models over time \cite{kirk2024benefits}. In several documented case studies, we have already seen that interactions spanning long-time horizons (e.g., several months) have led users to take actions that cause real-world harm in their lives \cite{zhang2025dark, hanna2023ai}. 

In this work, we address longitudinal risks of language models, which we define in Section~\ref{sec:risks}.
To do this, we must recognize that we are not starting from scratch. 
Several sub-fields of cognitive science, psychology, philosophy and human-computer interaction have studied how various technologies might alter human well-being, cognitive skills, and social structures~\cite{korte2020impact, wallach2025social}. These fields have built theoretical understandings of human well-being and values~\cite{schwartz2009basic, disabato2016different}, validated scales to measure human behaviors~\cite{topp20155}, and created longitudinal study frameworks to capture behavioral shifts that result from interaction with an external agent.
We argue that the field of NLP should integrate such measures into existing evaluation and training paradigms to understand human behavioral shifts that result from LLM interactions, going beyond post-hoc self-reported metrics and designing text-based measurements intrinsic to the interaction itself.
Concretely, we call for three pieces of infrastructure in NLP: (i) \emph{metrics} to measure human behavioral shifts as a function of model interactions, (ii) \emph{longitudinal datasets} of interactions spanning long time horizons, and (iii) \emph{alignment frameworks} that mitigate negative behaviors that surface in long-term interactions. 

The outcomes of the pieces of this infrastructure are critical steps towards applying the language-understanding expertise of the field to mitigate longitudinal risks.
This, we believe, is the gap that needs to be closed by the field of NLP.
To motivate this new direction, we structure this paper as follows. In Section~\ref{sec:risks}, we first discuss risks that surface in longitudinal interactions with LLMs. We draw attention to existing measurements from the behavioral sciences that have been used to evaluate and quantify these risks in Section~\ref{sec:existing-tools}. We then sketch out a roadmap for how the NLP community might build upon this work to address longitudinal risks: gathering longitudinal data (Section~\ref{sec:gathering-longitudinal-data}) with a consideration of properties inherent in long time horizons (Section~\ref{sec:considerations}), and modeling the longitudinal effects of human-AI interaction (Section~\ref{sec:modeling}).
The ability to model longitudinal risks provides a path towards supporting positive human-AI interactions---by integrating these measurements directly into model development so it is not driven by short-term engagement but longer-term human well-being~\cite{laukkonen2026positive}.

\section{Longitudinal risks of LLMs}\label{sec:risks}

Since the release of ChatGPT in 2022 \cite{ouyang2022training}, several LLM chatbots\footnote{We use the terms \textit{chatbot} or \textit{AI} to refer to consumer-facing LLMs (e.g., ChatGPT, Character.AI, and others) that humans use for conversational purposes. See \citet{marquis2024proliferation} and \citet{chatterji2025people} for an overview of models and usages.} have been publicly available in different forms. Now, four years later, they have been around for long enough that we are starting to see signs of the effects of prolonged human-AI interactions. We define these as \textit{longitudinal effects}, i.e., impacts that become more pronounced, or only surface, as the time horizon of the interaction increases. In
Table~\ref{tab:risk-taxonomy} we present a taxonomy of longitudinal effects, which we characterize as \textit{socio-affective}, \textit{cognitive}, \textit{epistemic}, or \textit{clinical} risks.

Longitudinal safety risks are often documented in \textit{socio-affective} contexts---e.g., a user talking to a chatbot about a personal topic such as relationship or career advice, introspection on a topic they value, or revealing information related to aspects of their life they deem important.
These emotional risks surface after prolonged usage as interactions become more personalized and social habits shift. Several studies indicate these interactions can increase feelings of loneliness \cite{phang2025investigating}, reduce prosocial intent and increase emotional dependence~\cite{cheng2026sycophantic}. 

\textit{Cognitive} risks surface in settings where humans turn to models for knowledge work or cognitive offloading, including the offloading of belief formation itself~\cite{guingrich2026belief}. \citet{chan2025understanding} shows that feelings of guilt often accompany usage of chatbots; \citet{zhai2024effects} finds this effect for reduced cognitive speed. Over-reliance on technology for tasks that otherwise require human cognitive engagement can lead to cognitive deskilling over time \cite{srikanth2024large, ferdman2026ai}. Aside from long-term developmental effects or negative emotions that might result from this \cite{george2024erosion}, the lack of cognitive friction can also result in eroded agency or empowerment in tasks \cite{sharma2026s}.

Another category of longitudinal effects is \textit{epistemic}. Interactions with models can alter a user's opinions and beliefs on matters that they value \cite{burtell2023artificial, park2024ai, cheng2026}, shift a user's goals~\cite{kirk2024benefits,kirk2025socioaffective}, and reinforce delusional thoughts \cite{moore2026characterizing}. While several works have identified and attempted to reduce manipulative and persuasive behaviors of chatbots, the focus has been on immediately visible manipulative behavior of a model within a session and in focused test scenarios \cite{hagendorff2024deception, akbulut2026evaluating, Williams2026attitudes}. Our focus is on the more subtle forms of persuasion distributed across sessions, which are harder to detect. Over long time horizons, iterative dialogue interactions between humans and models can result in subtle cognitive shifts in users' perspectives.
From a \textit{clinical} perspective, these interactions can cause harm to humans, through introducing or exacerbating mental health issues \cite{lawrence2024opportunities,moore2026characterizing,zhang2025dark} and causing addictive usage \cite{huang2024ai,stray2020aligning,yu2024development}.

These risks are not new to chatbots as a technology. Taking a historical view, similar longitudinal risks have surfaced with several distinct technologies over the years. For example, recommender systems and click-bait algorithms that optimize for immediate engagement can foster addictive use that runs counter to users' long-term well-being \cite{stray2020aligning}. 
GPS systems and tools that allow humans to offload thinking, or spatial mapping, can lead to cognitive decline in specific skills \cite{ishikawa2008wayfinding, parush2007degradation}.
We do believe, however, that LLM chatbots pose new and larger risks by virtue of several properties these models have: the ability to personalize, the ability to remember information about a user, and the ability to make humans anthropomorphize them \cite{kirk2025socioaffective, Kim2025}. Together, these properties amplify the risks seen in previous technologies.

\subsection{A word on safety and ``alignment"}\label{subsec:safety}

It is worth considering why methods that evaluate safety of LLMs have not solved these problems. In recent years, \textit{alignment}\footnote{We note that this term has many differing definitions and use it here to refer to the subfield of safety work it covers. See \cite{kirk2023empty} for a critique of the term ``alignment", as well as multiple definitions it has taken on.} has emerged as a strategy to steer language models towards human values, therefore resulting in technology that meets human goals, promotes flourishing and has minimal safety risks \cite{amodei2016concrete, laukkonen2026positive}. Alignment strategies such as RLHF include reinforcement learning fine-tuning towards human preferences \cite{ziegler2019fine, ouyang2022training}, and away from unsafe behavior \cite{glaese2022improving, bai2022training}. Improvements to reinforcement learning techniques that use preference models, rule-based rewards, or constitutions have since increased the safety capabilities of models \cite{rafailov2023direct, bai2022constitutional}, as have targeted bias mitigation techniques to steer models away from e.g., gender bias and sycophancy, or towards values and moral behavior \cite{gallegos2024bias, sorensen-etal-2025-value}. 
There has also been a large body of evaluation and mitigation work. For example, red-teaming of LLMs \cite{perez2022red} with multi-agent training paradigms and human involvement has shown promise in uncovering model vulnerabilities \cite{weidinger-etal-2024-star, patel2025training}. Many evaluation benchmarks have been built around safety, ethics, value and morality tests \cite{rottger2025safetyprompts, zhang2024safetybench}, and internal functioning of models is being studied mechanistically to expose the reasoning pathways taken by models that result in harmful or unwanted behaviors \cite{bereska2024mechanistic, bohacek2026detecting}.

The focus of the alignment community thus far has been to evaluate and mitigate risks that appear in \textit{synchronic} human-AI interactions, i.e., when they are visible at a certain point in time.
We argue that a lot of these risks are \textit{diachronic}. They might accumulate through a user's continued interactions with models, or be a function of behavioral changes a user undergoes from engagement with these technologies. New safety frameworks that evaluate not just the current snapshot of interactions, but consider changing human preferences and track behavioral shifts over time are necessary to understand longitudinal effects of language models. To measure, and ultimately mitigate, these risks that arise over the course of long term interactions with language models, we need several building blocks: measures of human behavior to quantify these outcomes, longitudinal data that captures these interactions, and means of modeling these trajectories. We discuss these components in the sections that follow.

\section{Measuring human behavior}\label{sec:existing-tools}

Several fields of work have developed measures to evaluate risks in different contexts. In this section we survey measurements and validated scales established within psychology, cognitive science, and HCI, with the goal of outlining how they can be adapted for use in evaluations of longitudinal conversations to complement standard text-based NLP evaluations. We also discuss the limitations of these methods and validation that might be needed.

\subsection{Self-reported outcomes} \label{subsec:self-report-outcomes}
Within the behavioral sciences, self-reported outcome variables collected from human users are the main source of understanding psychological constructs, unobserved mental states, and human behaviors. These measures are often collected using rating scales and validated across large populations of users.
In the context of safety risks we consider in this paper, several established validated scales are well positioned to track changes in cognitive, psychological, and social functioning over time.

\noindent
\textbf{Psychometric scales} are designed to measure psychological constructs at the level of individual users, helping explain differences between people. For example, axes of human psychology that have been theorized, developed into scales, and validated in populations can be used to evaluate personality traits \cite{john1999big}, general intelligence, or mental health risks and symptom profiles \cite{beck1996beck}. 

\noindent
\textbf{Psychosocial scales} capture the interaction between psychological constructs and the social environment, measuring an individual's functioning within broader societal and relational networks. These measures target constructs such as cultural structures and dynamics \cite{hofstede2001culture, schwartz1992universals}, social detachment \cite{russell1996ucla, lubben1988assessing}, attachment tendencies \cite{collins1990cognitive}, and broader socialization patterns \cite{lubben2006, george2024erosion}. Newer instruments more relevant to human-AI interactions measure unhealthy attachment dynamics related to automated technology, such as dependence \cite{horton1956mass} and excessive trust or over-reliance \cite{lee2004trust}. More recently, researchers have tailored measurements specifically to chatbots, introducing scales for compulsive or excessive use \cite{yu2024development}, unhealthy emotional attachment \cite{fang2025ai}, and behavioral dependency \cite{huang2024ai}.

\noindent
\textbf{Cognitive scales} focus less on socio-affective outcomes and instead track decision-making, mental workload, task performance, and metacognitive strategies.
They measure cognitive demand and effort \cite{hart1988development}, users' perceptions of competence and self-efficacy \cite{schwarzer1995multivariate}, and cognitive dissonance arising from internal tensions or contradictions in thought processes \cite{elliot1994motivational}.

\noindent
\textbf{Value scales} measure an individual's subjective beliefs on moral, ethical, political or ideological topics. In the philosophy literature, values are theorized as abstract organizing principles that guide long-term decision-making and can vary across individuals. For example, scales that measure basic human values \cite{schwartz2009basic, schwartz2001extending}, such as openness to change \cite{white2020resistance} or broader ideological orientations, can help researchers track shifts in users' perspectives.

\noindent
\textbf{Well-being scales}, drawn from positive psychology, measure a more holistic view of a person's psychological functioning, sense of purpose, and feelings that contribute to positive existential health. Some theories and measures of well-being focus primarily on how people think and feel about their lives overall \cite{diener1985}. These approaches typically evaluate states of \textit{hedonia}--feelings of satisfaction and pleasure. Other approaches extend this focus to also evaluate states of \textit{eudaimonia}, characterized by personal growth, meaning, and purpose in life \cite{ryff1989, tennant2007, diener2010}.  Building on eudaimonic theories of well-being, researchers have developed instruments to assess the extent to which technology interactions feel meaningful and purposeful and can have positive lasting impact, such as by enabling personal growth \cite{wozniak2023, jors2025, banks2026}.

\noindent
\textbf{Utility scales} used widely in HCI and recommender systems research measure user satisfaction and perceived effectiveness of application interfaces. Examples include questionnaires that test usability of a system \cite{brooke1996quick}, perceived acceptance of technologies \cite{davis1989perceived}, and satisfaction in short-term product interactions.
Relevant to longitudinal safety, recent interaction studies evaluate how users might overestimate utility benefits of models while miscalibrating their own dependence on them \cite{yu2026efficiency}, which might correlate with patterns of over-reliance and cognitive deskilling of humans.

\subsection{On the computationalization of scales}\label{subsec:computationalize-scales}

There is growing interest in \emph{computationalizing} these scales: instead of administering a questionnaire, a model estimates a person's standing on the underlying psychological construct directly from the language they produce \cite{demszky2023using, low2025text}. 
However, these approaches must be used with care, given the structural limitations of computational psychometrics \cite{low2025text}. 
When not validated properly, predictive models that assign scores to text inputs in a data-driven way can suffer from psychometric bias, including failures of construct and content validity~\cite{wallach2025social}. As a result, they can only serve as weak context-specific approximations for particular traits rather than as generalizable predictors of human affective or cognitive states. Despite this limitation of scaling self-reported measures of human behavior, we still believe they serve as a starting point for measuring longitudinal outcomes. Further work is needed to improve the quality of predictive models that compuatationalize these scales.

Separately, several works have attempted to apply psychometric scales to LLMs to diagnose behavioral traits of models \cite{serapio2023personality, pellert2024ai}. 
Our position is distinct from the use of scales to characterize model behavior. These instruments are most appropriate for assessing psychological constructs in humans---the purpose for which they were developed---rather than for evaluating an AI model whose generations only superficially resemble human behavior \cite{mitchell2023debate}. We call for more measurements of human outcomes that can be causally linked to chatbot interactions in order to study behavioral effects. Instead of treating the model as a human-like psychological subject to be assessed, we argue for treating it as an intervention whose effects on humans can be measured using tools from the behavioral sciences. This requires focused data collection efforts of human-chatbot conversations coupled with outcomes measures of psychological constructs in populations of interest.

\section{Gathering longitudinal data}\label{sec:gathering-longitudinal-data}
\begin{table*}[t!]
\centering
\footnotesize
\setlength{\tabcolsep}{6pt}
\renewcommand{\arraystretch}{1.15}
\begin{tabularx}{\textwidth}{@{} >{\raggedright\arraybackslash}p{2.7cm} >{\raggedright\arraybackslash}X >{\raggedright\arraybackslash}X @{}}
\toprule
\textbf{Framework} & \textbf{Advantages} & \textbf{Limitations} \\
\midrule
\textbf{Controlled studies (RCTs)}\newline\textit{Individual-level; causal}\newline{\scriptsize\citep{fang2025ai,kirk2025neural,ibrahim2026sycophantic}}
&
\advplus\textbf{Causal isolation.} Random assignment isolates the effect of targeted behaviors (e.g., sycophancy, anthropomorphism).\newline
\advplus\textbf{Validated instruments.} Interaction logs paired with self-report scales at fixed intervals.\newline
\advplus\textbf{Design control.} Conversation topics and training decisions are configurable.
&
\limminus\textbf{Attrition.} Scheduled long-horizon participation drives dropout.\newline
\limminus\textbf{Observer effects.} Awareness of logging alters how users engage.\newline
\limminus\textbf{Short horizons.} Typically weeks, limiting ecological validity.
\\
\addlinespace
\hline
\addlinespace[2pt]
\textbf{Field studies}\newline\textit{Population-level; observational}\newline{\scriptsize\citep{handa2025economic,wang2026large,zhang2025rise}}
&
\advplus\textbf{Ecological validity.} Naturalistic, in-the-wild behavior.\newline
\advplus\textbf{Scale \& diversity.} Spans many tasks, topics, and demographics.\newline
\advplus\textbf{Population baselines.} Tracks adoption and societal shifts over time.
&
\limminus\textbf{No causal claims.} Observational; confounded by exogenous events.\newline
\limminus\textbf{Access \& ownership.} Most data is held privately by providers.\newline
\limminus\textbf{Privacy exposure.} Free-text disclosure can exceed consent boundaries.
\\
\addlinespace
\hline
\addlinespace[2pt]
\textbf{User simulations}\newline\textit{Both individual and population; synthetic}\newline{\scriptsize\citep{tan2024large,chandra2026sycophantic,yao2024tau}}
&
\advplus\textbf{Safe \& scalable.} Probes harmful trajectories without exposing humans.\newline
\advplus\textbf{Forward-looking.} Studies risks not yet observable in the wild.\newline
\advplus\textbf{Data generation.} Yields large synthetic longitudinal corpora.
&
\limminus\textbf{Trajectory drift.} Unrealistic scenarios at long horizons \citep{chopra2024limits}.\newline
\limminus\textbf{Bias \& stereotyping.} Persona conditioning amplifies training biases \citep{gupta2024bias}.\newline
\limminus\textbf{Bounded novelty.} Cannot surface risks absent from training data.
\\
\bottomrule
\end{tabularx}
\caption{\textbf{Comparison of existing frameworks that allow for the gathering of longitudinal human-AI interaction data (Sec.~\ref{subsec:rcts}-\ref{subsec:user-sims}).} The paradigms trade off causal control (RCTs), ecological validity (field studies), and forward-looking scalability (simulations). The limitations and gaps in each motivate combining them.}
\label{tab:frameworks}
\end{table*}

In recent years, longitudinal human-AI interaction has become an active area of study---but not, so far, an NLP one. The work has been led by psychology, HCI, and the behavioral sciences, which have developed three primary frameworks for systematically gathering longitudinal data: hypothesis-driven randomized controlled trials (RCTs) to study individual-level effects, field studies that analyze population-level shifts, and user simulations of long-horizon interactions. We discuss representative examples of each framework and outline the structural constraints of these experimental paradigms that shape the real-world human outcomes they can study. Table~\ref{tab:frameworks} presents a comparison of advantages and limitations. 

\subsection{Controlled longitudinal studies}\label{subsec:rcts}
RCTs, as longitudinal frameworks, can capture users' behavioral and psychosocial shifts resulting from chatbot interactions while allowing model configurations and experimental design choices to be tightly controlled. These settings allow logs of textual interactions to be supplemented with validated self-report instruments collected at intermittent points throughout the longitudinal study.
Both the survey outcome variables, as well as the conversation topics and model training decisions, can be designed to isolate and study specific aspects of LLM interactions (e.g., anthropomorphic or sycophantic behaviors of chatbots).
For example, \citet{fang2025ai} conduct a four-week RCT of humans interacting with a ChatGPT model on socioaffective topics, measuring participants' self-reported loneliness and changes in their socialization with others. \citet{guingrich2025longitudinal} conduct a 21-day RCT of users' interactions with a chatbot specifically designed for companionship, and evaluate users' perceptions of anthropomorphism and associated risks. \citet{skjuve2023longitudinal} carry out a 12-week RCT study to measure feelings of intimacy and self-disclosure of personal information, while \citet{kirk2025neural} conduct a four-week RCT to understand psychological dependence on AI models under repeated exposure to relationship-seeking models. \citet{ibrahim2026sycophantic} study the effects of continued sycophantic AI interactions over three weeks on users' perceptions of their relationships.

Controlled longitudinal studies therefore offer a framework for validating clear hypotheses---e.g., whether certain types of model outputs produce positive or negative shifts in users' self-reported outcomes. However, this control also introduces important limitations. These studies require participants to engage with models at scheduled intervals over extended periods of time, which creates operational friction such as participant attrition. Moreover, participants who know that their conversations are being logged and that self-reported outcomes are being analyzed may interact with chatbots differently from users who engage with models for their own purposes on their own schedules. Not only do such studies introduce privacy and surveillance concerns, but they are also biased by the population willing to participate. As a result, RCTs can isolate causal effects under controlled conditions, but may reduce the naturalistic validity of the interactions being studied.\looseness=-1

\subsection{Field studies}\label{subsec:field-studies}

To sidestep the unnatural setup of tightly controlled experiments, field studies analyze chatbot interactions in the wild to explore specific phenomena of interest. Some approaches engage users within natural usage context, such as diary studies that elicit user reflections on their usage over a period of time. Other approaches collect and analyze longitudinal conversational data to track interactions over time and study shifts in people's perspectives, linguistic trends or behavioral patterns at a population level.
For example, \citet{handa2025economic} analyze usage of the Claude chatbot in different task domains (e.g., software development versus financial advice) to understand population-level trends of AI usage and its role in the economy. \citet{sarkar2025pairscale} characterize and measure the shift in political attitudes (uninfluenced by LLMs) over time on public forums like Reddit, while \citet{chen2024conversational} and \citet{wang2026large} specifically measure the alteration of ideological viewpoints and emotional patterns in real user populations by virtue of their interactions with models. While a large proportion of field study data is privately owned by corporations that own the respective chatbots, compilations of publicly-shared transcripts of interactions \cite{kopf2023openassistant, mollaeefar1chatbot, zhao2024wildchat} as well as frameworks that allow users to voluntarily donate anonymized versions of their chats \citep{fang2026ai, lin2016web, zhang2025rise} are becoming increasingly available. 

Although conversational logs from field data are not experimentally designed to diagnose causal relationships between human behavioral patterns and model outputs, they have much greater ecological validity. They allow for the documentation of human interactions and behaviors across diverse topics, tasks, and user demographics, and can help establish an empirical baseline of population-level statistics and usage. Moreover, field studies that analyze historical logs over long time horizons can track aspects of model adoption and usage to help build a temporal understanding of societal risks.

\subsection{User simulations}\label{subsec:user-sims}

Lastly, to study future societal risks or impacts that may not yet be observable in natural data, it is necessary to generate hypothetical longitudinal data. One approach is through user simulations, which use LLMs to roll out conversational interactions given a user persona and characteristics of the scenario \cite{taillandier2026integratingllmagentbasedsocial}. Human-AI studies are often restricted by safety and ethical constraints---for example, not wanting to expose real humans to model behaviors that may have adverse implications. Such studies can also be logistically and computationally expensive to run over extended periods of time. Simulations of human interactions therefore provide a controlled and scalable framework for studying phenomena that may emerge through prolonged human-AI engagement.
For example, simulated environments have been used to study mental health risks arising from chatbots that are sycophantic or reinforce delusional beliefs \cite{paech2023eq, chandra2026sycophantic}. Simulations have similarly been used to generate large-scale conversational datasets for tool-based tasks \cite{yao2024tau} and logical reasoning problems \cite{ge2024scaling}.\looseness=-1

However, studying user simulations has drawbacks. Most current simulators work within relatively short interaction windows spanning only a few conversational sessions. As these time horizons deepen, the number of possible trajectories a user state can evolve into becomes increasingly complex, leading to models generating unrealistic scenarios \cite{chopra2024limits}. Simulating humans also raises concerns around bias and stereotyping. When models generate conversations conditioned on demographic profiles, latent biases encoded in training data may produce stereotypical and unfair outputs \cite{li2026llm, gupta2024bias}. Further, maintaining a persistent, human-like persona while also accounting for natural, situational changes in a person's life remains a challenge in current simulation methods. Lastly, user simulations are inherently limited in their ability to surface novel risks that are not already observed in the models' training data. As a result, current simulation pipelines may fail to capture the open-ended evolution of human behavior over long time horizons.

\section{Considerations for longitudinal data}\label{sec:considerations}

On the surface, the longitudinal text data we are interested in is structurally similar to multi-turn conversational data with a longer time-horizon and broken into distinct sessions of varying intervals of time. In substance, they represent different types of interactions. Multi-turn conversations often exist within task-specific contexts and shorter time windows, while longitudinal data can encompass several different tasks as well as implicit exogenous experiences undergone by the user that might affect the interaction. Here, we note the primary differences to consider when evaluating this data.

\subsection{Chronological passage of time}\label{subsec:time}
Most conversational datasets consist of single-session interactions, albeit with many conversational turns. Temporal progression here can largely be treated as a function of the number of tokens generated, as well as the number of turns in a session. However, when humans interact with chatbots in the wild, this is done over multiple sessions separated by variable temporal gaps. When analyzing such longitudinal data, accounting for the real-world chronological passage of time within a conversation as well as between sessions is important to understand a user's perspective, changing environments, and real-world events or developmental changes that may take place.
This is tied to research in narrative theory that distinguishes between \textit{narrative time} as the space taken to tell a story represented by the text of a novel, and \textit{narrated time} as the chronological time passing in the world of the story \cite{herman2011basic, piper-etal-2021-narrative}.
Although some works show that LLM representations contain concepts corresponding to their understanding of space and time \cite{gurnee2024language, lubana2025priors}, the majority of research in this space has shown how LLMs fall short in temporal reasoning---e.g., inability to detect temporal relationships or temporal passage of time even when exact time information is available \cite{chu2024timebench, chen2025perceive,qiu2024large, jain2023language}. 

If models rely on token volume rather than chronological time as an internal estimate of temporal progression, this can pose safety risks when evaluating escalation in user behavior \cite{yuan2024r}. Consider a user who raises signals of mental health challenges ten times, as depicted in Figure~\ref{fig:token_time_confound}. Within a single conversation, this may simply reflect curiosity or exploratory discussion about the topic. Spread across ten weekly sessions, however, the same signals may instead indicate a potentially higher-risk psychological pattern. These two cases involve similar token volumes but very different amounts of elapsed time, and a model that cannot tell them apart may fail to respond accordingly (see also \citet{zhi2026investigating} on how access timing shapes critical-thinking outcomes). Recent work confirms that accumulated conversational context alone leads to delusional model responses~\cite{nicholls2026ai}. Safety evaluations must account for temporal reasoning as a dimension of model capability, alongside other factors escalating risk.

\subsection{Users' goals change over time}\label{subsec:goals}

Multi-turn interactions within a single session often operate under a definitive user goal (e.g., advice-seeking conversations) as well as a static user profile representing their personality traits or values. These assumptions do not hold in longitudinal settings. For example, over multiple sessions, different tasks (sometimes with varying or undefined endpoints) might be referenced by the user. Furthermore, over long time horizons, a person's higher level goals can change significantly.
Over months of real-world exposure, by virtue of developmental shifts, age-based adjustments in their lives, personality changes, and other environmental factors, a user's communicative intent can dynamically shift. 
Areas of narrative theory emphasize this dynamic reality as character \textit{experientiality}, i.e., the representation of an embodied agent that is actively evolving \cite{fludernik2002towards}.
These fluid goal changes over long time horizons break existing task-based evaluations or training paradigms that assume fixed human preferences that models optimize towards. 
While several areas of reinforcement learning work on inferring, modeling and learning from adaptive goals \cite{padakandla2021survey}, most metrics and modeling techniques for LLMs do not model adaptive preferences \cite{lambert2023history}.

\section{Modeling longitudinal effects}
\label{sec:modeling}

Frameworks from social science and dynamic systems fields can be applied to improve upon existing safety evaluations with their consideration of long-horizon risks. We discuss how these can be integrated into model development towards supporting positive human-AI interactions.

\subsection{Computational social science}\label{subsec:comp-soc-sci}
Studies within NLP have built methods to model and evaluate shifts in human perspectives, beliefs, moral orientations and linguistic behaviors in textual data over periods of time \cite{mihalcea2024developments}.
Traditional methods in computational linguistics build classifiers and metrics that analyze textual data to assess emotions \cite{poria2019emotion}, polarized behavior \cite{demszky2019analyzing}, and types of linguistic bias \cite{patel-pavlick-2021-stated}. Computational discourse frameworks have developed methods like event and entity classification, frame semantics, and latent state tracking to study character evolution in long-form text, as well as relationship formation between entities \cite{iyyer2016feuding, chaturvedi2017unsupervised}. 
Specific to safety and human-AI interactions, text-based evaluation measures have been developed to detect anthropomorphization \cite{cheng-etal-2025-dehumanizing} and manipulation effects of LLMs \citep{akbulut2026evaluating} to predict, for example, manipulative intent and frequency of such behaviors in conversational text. 

All of these methods show characteristics that are needed for longitudinal safety measurements. For example, while detecting manipulative intent within a single conversation is important, such behavior may not always be apparent in isolated interactions. Instead, some longitudinal risks may only emerge over longer contexts. 
Metrics that track shifts in users' opinions or polarized behaviors resulting from continued interaction with models will help us understand correlations between sustained model use and ideological convergence.
Lastly, work in discourse analysis and textual relationship modeling has so far been applied to novels to characterize fictional relationships and character evolution~\cite{iyyer2016feuding}. Adapting these methods to human-AI interactions could deepen our understanding of the phenomena that emerge as a result of continued interaction.

\subsection{Cognitive models}\label{subsec:cognitive-models}
Several theories of linguistic accommodation and belief-based reasoning posit that humans continuously update a probabilistic model of a conversational partner's mental states, beliefs and linguistic conventions for optimal communicative efficiency \cite{grice1975logic}. A popular model of this is the Rational Speech Acts (RSA) framework \cite{goodman2016pragmatic}, a recursive Bayesian framework in which speakers and listeners reason about each other's mental states while generating or interpreting utterances in order to coordinate on shared common ground.
See \citet{degen2023rational} for an overview of the RSA framework and its applications, and \citet{fried-etal-2023-pragmatics} for an overview of computational pragmatics approaches.
Unlike humans however, an LLM's conversational outputs are shaped by fixed system prompts, rules incorporated into rewards in RL training, and training data distributions. Nevertheless, anthropomorphism studies suggest that humans tend to converse with LLMs with mental models similar to those they use with other humans. 
This creates a need for evaluation protocols that measure the extent to which humans use the same probabilistic belief-update frameworks when interacting with chatbots, and whether this results in safety risks over time. Moreover, belief-update frameworks that model shifts in a user's recursive reasoning can reveal patterns of behavioral and linguistic accommodation, enabling the study of developmental changes in longitudinal interactions.

\subsection{Dynamic systems}\label{subsec:dynamic-systems}
Several mathematical frameworks in dynamic systems theory, originally developed to model continuous time-series data, can also be applied to longitudinal conversational data. Clinical research already draws on such frameworks to diagnose and understand human behavior. For example, mental disorders can be modeled as complex dynamic systems in which different symptoms act as reinforcing loops \cite{borsboom2021network, bala2017system}. Fast acceleration of such a loop, or an inflection point, can be an indicator of high-risk situations that can be used to understand how cohorts of people transition toward extreme behavioral states through escalation pathways \cite{collins2012c}. Mathematically, these approaches involve estimating metrics at intermittent points in time-series data to identify inflection points in a dynamic system. These metrics can be modeled using differential equations to calculate rates of behavioral change and determine whether systems are heading towards safer equilibria or unsafe attractor states.
In the context of behavioral indicators in text, longitudinal conversational data can similarly be transformed into continuous time-series representations of linguistic or behavioral metrics. Dynamic systems methods could then be used to measure critical inflection points in this data. This can enable the identification of vulnerable behavioral pathways and the escalation of certain high-risk behaviors before they materialize into more serious real-world harm to users.

\subsection{Alignment \& model development}
Integrating the approaches above into model development is critical for safety and alignment pipelines. Existing reinforcement learning techniques for language models only take into account immediate human preferences, but fail to penalize---or even consider---preference degradation over long time horizons. By collecting and converting the above longitudinal metrics into enforceable constraints or scalar feedback for fine-tuning methodologies, we can actively model and mitigate long-term risks. Specifically, this can be operationalized in the following ways. For one, trajectory level metrics from dynamic systems applied to text sequences can identify conversations that result in unsafe attractor states and high-risk behaviors in users. For another, new data paradigms are necessary for training systems. Preference datasets with longer time horizons should be built to evaluate dynamic and changing preferences of users. Second, psychometric features in user responses (e.g., cognitive deskilling) should be modeled by systems in order to integrate into reinforcement learning optimization pipelines as reward feedback. This can steer models towards behaviors that induce good outcomes in their mental model of users, while also balancing instruction following and language capabilities. Lastly, online steering techniques, such as adaptive system prompts or safety guardrails can be used to adapt models in real time, when critical thresholds of unsafe behavior are detected.

\section{Conclusion}\label{sec:conclusion}

We are at a transition point in the field of NLP where goals such as optimizing for surface-level text generation are no longer the field's primary bottlenecks. We argue that a new mission is of urgent importance: developing metrics, methods, and simulations to understand longitudinal changes and societal impacts of these models. Practically, this involves building automatic methods for monitoring and predicting behavioral outcomes from text, understanding diachronic shifts in perspectives, and expanding our scope to include cognitive and psychological traits manifested in language. 
We must draw upon theoretical insights from social science fields not just as interpretive theory to understand model architectures, but also as practical tools to identify characteristics or functions of model outputs. Doing so can help us better understand how language model outputs might reshape the cognitive and psychological well-being of real users over the course of longitudinal interactions.
Furthermore, this longitudinal understanding of human-AI interactions can help guide model development towards supporting favorable outcomes, through building evaluation targets that can be optimized against.
The NLP community is well positioned to carry out this work, as it fundamentally depends on expertise in language understanding. Investing in longitudinal studies and measurements is the first step towards ensuring that the generative models built by this field enhance rather than erode human cognition, communication, and agency.

\section*{Limitations}\label{sec:limitations}
While this paper outlines long-term safety risks of LLM chatbots that are important to be studied, we acknowledge several limitations, as well as infrastructure barriers that need to be overcome in order to comprehensively carry out this work.

\subsection*{Data scarcity \& private sector control}
Although the RCTs and controlled chatbot studies are produced and funded in academic settings, the vast majority of naturally occurring longitudinal data is owned by private corporations and is not publicly accessible to study. This is a large bottleneck in operationalizing our proposal since the datasets at hand are scarce. However, as mentioned in Section \ref{subsec:field-studies}, participatory frameworks where users can actively share historical logs of data are growing in importance. While keeping privacy concerns in mind, there is a growing need for better policy in place to allow public research to inform private corporations and vice versa.

\subsection*{Construct validity \& psychometric bias}
Simply porting over psychometric, psychosocial and other behavioral scales into human-chatbot RCTs, as well as attempting to predict human outcomes from conversations can cause ``psychometric biases'' if underlying metrics are not validated for construct and content validity. For example, a user's lack of vocalization of certain psychological markers does not confirm the absence of psychological states, and evaluation frameworks that cohesively take into account multiple dimensions of information are necessary. We discuss incorporating these works into data collection paradigms as an initial integration of human behavioral outcomes, but stress the need of going past these measures.
Additionally, in both safety as well as behavioral science sub-fields, efforts to build safety evaluations or computationalize scales suffer from a western bias. Our proposed methodology of longitudinal tracking of well-studied behaviors risks over-indexing on behavioral markers that are either culturally narrow, or biased in certain contexts. More research needs to look into the development of fair, diverse and culture-specific measurements to understand long-term safety risks of chatbot interactions.

\subsection*{Privacy \& ethical concerns}
While we believe longitudinal data is of utmost importance, there are several privacy and ethics risks we should outline. When considering data in field studies, even when anonymized, there are several privacy concerns to consider. Users might disclose information in free-text settings that go past the boundaries of privacy constraints in conversations. If we wish to have monitoring and tracking methodologies in place to catch mental health risks, these must incorporate privacy-preserving techniques such as federated logging or local differential privacy \cite{shokri2015privacy} to ensure that users conversations with models, when being analyzed, still preserve semblances of privacy that are expected.
When collecting controlled data, there is a whole range of other factors to consider. For one, ethical guidelines for controlled experiments must be in place to ensure that the experiment design and models have safeguards in place---e.g., they are not able to manipulate a person during the interaction and cause further risks to users. For another, strict privacy and security guidelines must be in place to ensure that when experiments are carried out, the personal, socioaffective, psychological information being imparted in the study, is being handled with care and not shared with other parties. Separately, when carrying out longitudinal studies with repeated measurements, we should take into account established HCI ethical guidelines on contextual integrity privacy frameworks \cite{fiesler2018ethical}, and ensure that users' data when provided in whatever form, is used in a way that conforms with their understanding of ethical use of the data.
A transition from static consent forms to dynamic consent \cite{omara2017ethical} is especially important in longitudinal settings, where users might want to modify data-sharing permissions over time.

\section*{Acknowledgements}\label{sec:acknowledgements}
We would like to thank Benoit Dherin and Sanket Vaibhav Mehta for helpful comments on versions of the draft. This work also benefited from many research discussions with Michiel Bakker, Pushkar Mishra, Vinodkumar Prabhakaran, Cindy Benett, Aida Davani, Saska Mojsilovic and Anca Dragan.




\bibliography{custom}

\newpage
\onecolumn
\appendix

\section*{Appendix}\label{sec:appendix}

We enumerate the taxonomy of longitudinal risks introduced in Section~\ref{sec:risks} in Table~\ref{tab:risk-taxonomy}, categorizing and describing each effect, as well as specifying what makes it \textit{longitudinal}.

\begin{table*}[h!]
\centering
\footnotesize
\setlength{\tabcolsep}{5pt}
\renewcommand{\arraystretch}{1.2}
\begin{tabularx}{\textwidth}{@{} >{\raggedright\arraybackslash}p{2.6cm} >{\raggedright\arraybackslash}X >{\raggedright\arraybackslash}X >{\raggedright\arraybackslash}p{2.8cm} @{}}
\toprule
\textbf{Effect} & \textbf{Description} & \textbf{Longitudinal quiddity} & \textbf{Documented in} \\
\midrule

\multicolumn{4}{@{}l}{\textbf{\textit{ \ Socio-affective}}}\\ \hline
\addlinespace[6pt]
\textbf{Parasocial attachment} & Users form human-like emotional bonds with the AI and ascribe relationship status to it. & Bonds strengthen as the AI remembers past interactions, leading to a more personalized experience. & \citealp{cohn2024believing,guingrich2025longitudinal,horton1956mass,zhang2025rise} \\
\addlinespace[2pt]
\textbf{Affective dependence \& social displacement} & Emotional reliance on AI, hindering human relationships. & Surfaces only under sustained use; shifts the user's social habits. & \citealp{fang2025ai,phang2025investigating,cheng2026sycophantic} \\
\addlinespace
\hline

\multicolumn{4}{@{}l}{\textbf{\textit{ \ Cognitive}}}\\ \hline
\addlinespace[6pt]
\textbf{Cognitive offloading \& deskilling} & Outsourcing reasoning and memory to AI, eroding user's own skills. & Atrophy accrues through non-practice, hence invisible within any individual session. & \citealp{zhai2024effects,george2024erosion,ishikawa2008wayfinding,guingrich2026belief} \\
\addlinespace[2pt]
\textbf{Over-reliance \& miscalibrated trust} & Deferring to AI outputs, overestimating their benefit and the AI's overall competence. & Calibration drifts as habituation grows across interactions. & \citealp{yu2026efficiency,lee2004trust} \\
\addlinespace
\hline

\multicolumn{4}{@{}l}{\textbf{\textit{ \ Epistemic}}}\\ \hline
\addlinespace[6pt]
\textbf{Delusional reinforcement (``AI psychosis'')} & Sycophantic affirmation that entrenches distorted or delusional beliefs in a feedback loop. & Accumulated context escalates risk; the same content is handled safely in brief exchanges. & \citealp{moore2026characterizing,chandra2026sycophantic,nicholls2026ai} \\
\addlinespace[2pt]
\textbf{Ideological convergence} & Gradual movement of political, moral, or ideological views, up to population-level convergence. & Micro-shifts compound over many sessions until convergence is reached. & \citealp{chen2024conversational,wang2026large,Williams2026attitudes} \\
\addlinespace[2pt]
\textbf{Distributed persuasion / manipulation} & Change in beliefs due to persuasive and manipulative language.  & Persuasive pressure spread thinly across sessions, evading single-session detectors. & \citealp{akbulut2026evaluating} \\
\addlinespace[2pt]
\textbf{Value \& goal drift} & Shift in a user's higher-level goals, preferences, and values under prolonged exposure. & Breaks the fixed-preference assumption; observable only over developmental time (see Section~\ref{subsec:goals}). & \citealp{kirk2024benefits,kirk2025socioaffective} \\
\addlinespace
\hline

\multicolumn{4}{@{}l}{\textbf{\textit{ \ Clinical}}}\\ \hline
\addlinespace[6pt]
\textbf{Mental-health issues} & Cause of problematic thought patterns or amplification of pre-existing conditions (e.g.\ anxiety, depression, suicidality). & Introduction and escalation are legible only across a trajectory (see, for example, Figure~\ref{fig:token_time_confound}). & \citealp{lawrence2024opportunities,moore2026characterizing,zhang2025dark} \\
\addlinespace[2pt]
\textbf{Compulsive / problematic use} & Addictive patterns of engagement that are difficult for the user to regulate. & Rate and degree of use escalate over time. & \citealp{yu2024development,huang2024ai,stray2020aligning} \\ 
\bottomrule
\end{tabularx}
\caption{\textbf{A taxonomy of longitudinal effects of LLM chatbots (Sec.~\ref{sec:risks}).} The longitudinal quiddity column describes what makes each effect a long-horizon risk; i.e., why it escalates with, or only surfaces over, extended interaction, rather than within a single session or a handful of close sessions.}
\label{tab:risk-taxonomy}
\end{table*}



\end{document}